\documentclass[11pt]{article}
\usepackage{acl}

\usepackage{times}
\usepackage{latexsym}

\usepackage[T1]{fontenc}

\usepackage[utf8]{inputenc}

\usepackage{microtype}

\usepackage{inconsolata}
\usepackage{subcaption}
\usepackage{graphicx}
\usepackage{tabularx}
\usepackage{listings}
\usepackage{xcolor}
\usepackage{url}
\definecolor{codeblue}{RGB}{33,74,135}
\definecolor{codegray}{RGB}{120,120,120}
\definecolor{codepurple}{RGB}{122,17,88}
\definecolor{backgray}{RGB}{248,248,248}

\lstdefinelanguage{yaml}{
  sensitive=false,
  morecomment=[l]{\#},
  morestring=[b]',
  morestring=[b]",
  alsoletter={-},
  morekeywords={true,false,null,yes,no,on,off}
}

\usepackage{multirow}
\usepackage{booktabs}
\usepackage{pifont}           
\newcommand{\ck}{\ding{51}}   
\newcommand{\parti}{$\circ$}  
\newcommand{\na}{--}  
\title{DreamReader: An Interpretability Toolkit for Text-to-Image Models}

\author{
\textbf{Nirmalendu Prakash}\textsuperscript{1},
\textbf{Narmeen Oozeer}\textsuperscript{2},
\textbf{Michael Lan}\textsuperscript{2},
\textbf{Luka Samkharadze}\textsuperscript{2},\\
\textbf{Phillip Howard}\textsuperscript{3},
\textbf{Roy Ka-Wei Lee}\textsuperscript{1},
\textbf{Dhruv Nathawani}\textsuperscript{4},
\textbf{Shivam Raval}\textsuperscript{5},
\textbf{Amirali Abdullah}\textsuperscript{2,3} \\
\textsuperscript{1}Singapore University of Technology and\\
\hspace*{1.5em}Design \quad
\textsuperscript{2}Martian \quad
\textsuperscript{3}Thoughtworks \quad
\textsuperscript{4}NVIDIA \quad
\textsuperscript{5}Harvard University
\\[0.5em]
\small{
   \textbf{Correspondence:}
   \href{mailto:nirmalendu_prakash@mymail.sutd.edu.sg}{nirmalendu\_prakash@mymail.sutd.edu.sg}
}
}

\begin{document}
\maketitle
\begin{abstract}

Despite the rapid adoption of text-to-image (T2I) diffusion models, causal and representation-level analysis remains fragmented and largely limited to isolated probing techniques. To address this gap, we introduce DreamReader: a unified framework that formalizes diffusion interpretability as composable representation operators spanning activation extraction, causal patching, structured ablations, and activation steering across modules and timesteps. DreamReader provides a model-agnostic abstraction layer enabling systematic analysis and intervention across diffusion architectures. Beyond consolidating existing methods, DreamReader introduces three novel intervention primitives for diffusion models: (1) representation fine-tuning (LoReFT) for subspace-constrained internal adaptation; (2) classifier-guided gradient steering using MLP probes trained on activations; and (3) component-level cross-model mapping for systematic study of transferability of representations across modalities. These mechanisms allows us to do lightweight white-box interventions on T2I models by drawing inspiration from interpretability techniques on LLMs.

We demonstrate \textsc{DreamReader} through controlled experiments that (i) perform \emph{activation stitching} between two models, and (ii) apply \emph{LoReFT} to steer \emph{multiple activation units}, reliably injecting a target concept into the generated images. Experiments are specified declaratively and executed in controlled batched pipelines to enable reproducible large-scale analysis. Across multiple case studies, we show that techniques adapted from language model interpretability yield promising and controllable interventions in diffusion models. DreamReader \footnote{\url{https://github.com/Social-AI-Studio/T2I_Interp_toolkit}} is released as an open source toolkit for advancing research on T2I interpretability.

Video demo: \url{https://youtu.be/VIuacdjSMcU} \\
Live demo: \url{https://huggingface.co/spaces/withmartian/dream_reader_demo}

\end{abstract}
\section{Introduction}

Text-to-image (T2I) diffusion models can produce dramatically different outputs from minimal prompt changes, yet the internal mechanisms driving this behavior remain poorly understood. Prompt-based control is limited \cite{prompt_1} and such systems can exhibit biased, hateful, or otherwise undesirable behaviors under certain inputs \citep{unsafe_1, unsafe_2}. Ensuring reliable audit and control therefore requires systematic access to latents. However, interpretability methods for diffusion systems remain fragmented and architecture-specific, which makes it difficult to transfer state-of-the-art language-model interpretability methods to T2I settings.

Unfortunately this fragmentation is structural in nature. 
Unlike language models, diffusion systems are multi-stage generative stacks comprising text encoders, iterative denoising modules, decoders, and guidance mechanisms. Interventions can occur across timesteps, components, and representation types, creating a much larger intervention surface than in transformers.

\begin{figure*}[t]
\centering
\IfFileExists{images/workflow.pdf}{%
  \includegraphics[width=1\textwidth]{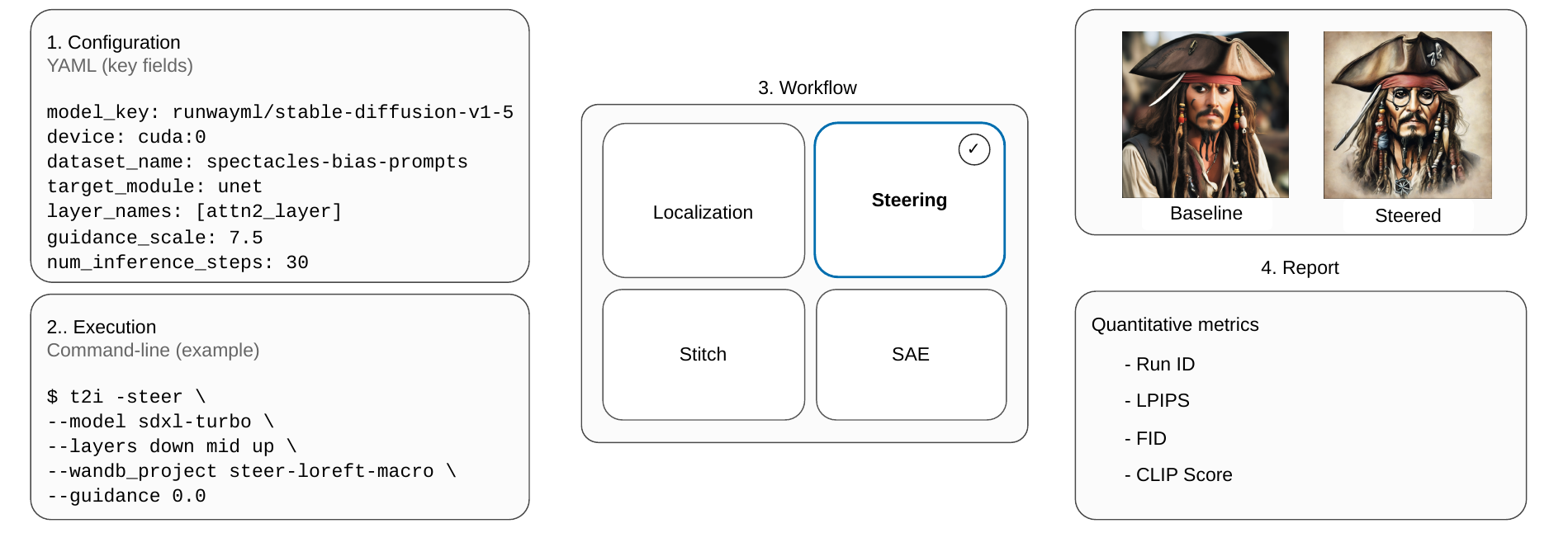}%
}{%
  \fbox{\parbox[c][4.5cm][c]{0.95\textwidth}{\centering Placeholder: workflow.pdf not bundled}}%
}

\caption{\textsc{DreamReader} provides a unified abstraction of the diffusion stack, exposing four core interpretability operators—Localization, Steering, Stitching, and Sparse Autoencoders (SAEs)—that can be composed within a shared interface. Shown above are the 4 steps in the analysis (shown here is an example for `Steering').
}
\label{fig:t2i-interp-workflows}
\end{figure*}

\begin{table*}[t]
\centering
\small

\begin{minipage}{\textwidth}
\vspace{2pt}
\setlength{\fboxsep}{6pt}
\noindent\fbox{%
\begin{minipage}{0.985\textwidth}
\textbf{Common across all workflows.}
\textbf{Inputs:} model identifier \& revision, dataset version/splits, prompts/batches;
\textbf{Hook sites:} dotted module paths (layer/block/head/timestep);
\textbf{Run control:} seeds, devices, mixed precision, logging/checkpointing.
\quad
\textbf{Common metrics:} CLIP similarity/alignment, LPIPS/SSIM, diversity summaries, runtime/memory.
\quad
\textbf{Common outputs:} generation grids, run metadata, config snapshots, checkpoints, and aggregated reports.
\end{minipage}}
\vspace{6pt}
\end{minipage}

\begin{tabularx}{\textwidth}{lXX}
\toprule
\textbf{Workflow} &
\textbf{Unique configuration / interventions} &
\textbf{Workflow-specific metrics \& outputs} \\
\midrule
\textbf{Localization} &
Intervention type (scaling/zeroing/corruption), strength, and timestep schedule. &
Effect ranking over sites/timesteps; influence reports (e.g. which hook sites are more causal). \\

\textbf{Steering} &
Method (CAA/K-Steer/LoREFT), learned direction(s), application site(s), scale(s), schedule(s). &
Steering success curves vs.\ strength/location; reusable steering vectors and ranking plots. \\

\textbf{Stitching} &
Mapper family (affine/MLP), training schedule/loss; module swapping and transfer settings. &
Compatibility heatmaps (i.e. cross-layer comparison between modules); stitched vs.\ native comparison grids; mapper checkpoints. \\

\textbf{SAEs} &
Dictionary size, sparsity/Top-$k$, activation source (raw/residual); feature editing scale. &
Dead-feature rate, sparsity stats; feature cards with top activations. \\
\bottomrule
\end{tabularx}

\caption{\textsc{DreamReader} workflows. Shared inputs/metrics/outputs are listed in the common block; each row reports only workflow-specific configuration knobs and deliverables. Technical terms are defined in Table~\ref{tab:workflow-glossary}.}
\label{tab:workflow-comparison}
\end{table*}

\begin{figure*}[t]
\centering
\IfFileExists{images/LoREFT_results.png}{%
  \includegraphics[width=1\textwidth]{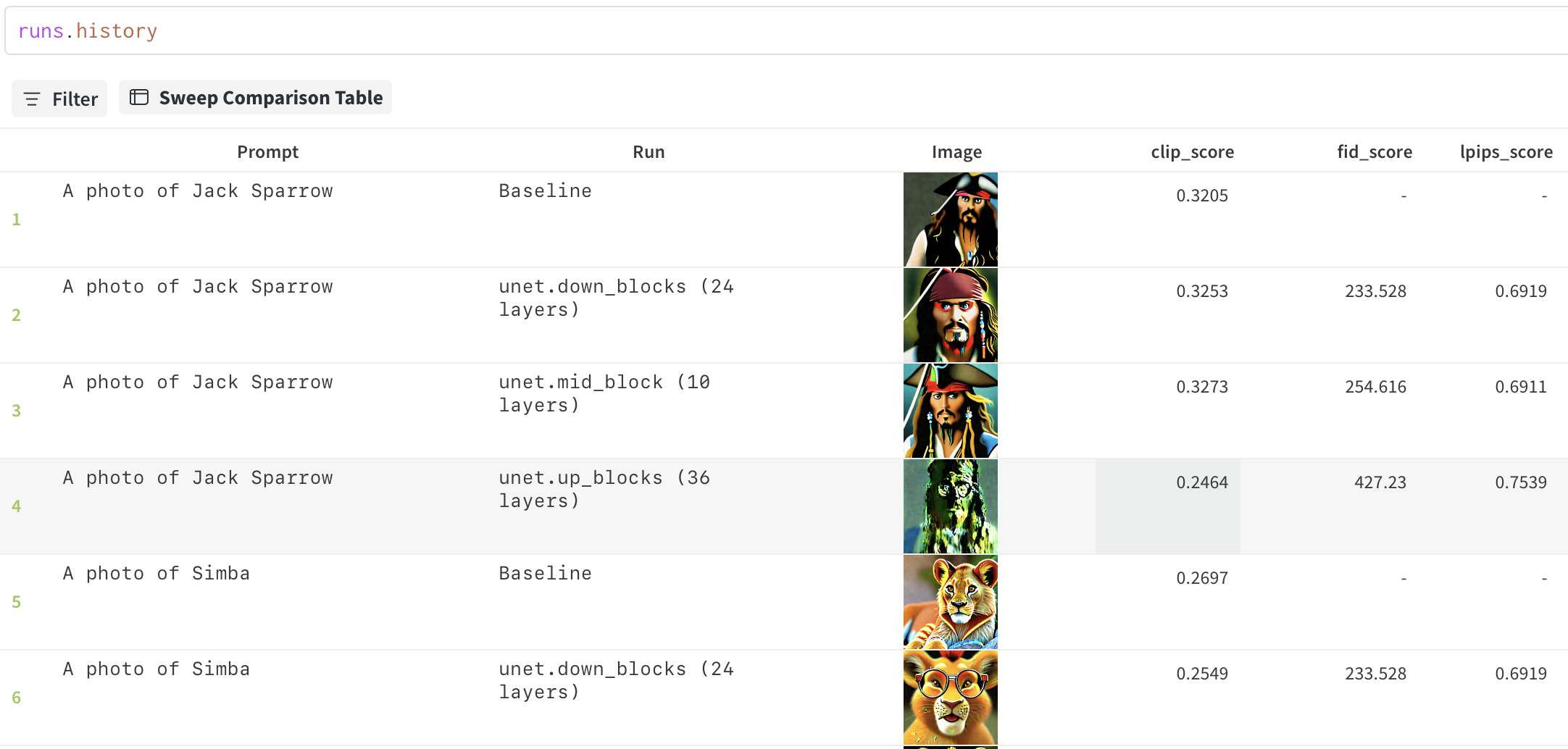}%
}{%
  \fbox{\parbox[c][4.5cm][c]{0.95\textwidth}{\centering Placeholder: LoREFT\_results.png not bundled}}%
}
\caption{LoREFT steering results on sample prompts for SDXL-Turbo. We steer the model to add spectacles by training a LoREFT module on cross-attention activations from different U-Net regions (down, mid, and up blocks). Steering effectiveness varies by prompt: for \textit{Simba}, down-block steering succeeds, whereas for \textit{Jack Sparrow}, up-block steering performs best. CLIP score measures prompt--image alignment; FID and LPIPS quantify the deviation from the baseline (unsteered) output. Results are logged and visualized with W\&B.}
\label{fig:loreft_result}
\end{figure*}
Consequently, existing diffusion interpretability research often targets isolated components rather than the diffusion stack as a whole. For example, Diffusion Lens probes intermediate representations of the text encoder across timesteps to analyze how semantic concepts evolve during denoising \citep{toker2024diffusion}. CASteer performs cross-attention head steering by applying concept-specific steering vectors to attention activations to suppress or manipulate visual concepts during generation \citep{gaintseva2025casteer}. Patch Explorer enables inspection and manipulation of individual cross-attention heads, visualizing how token-level conditioning influences spatial regions of the generated image \citep{grabe2025patch}. These tools provide valuable, component-specific insights, but typically operate at fixed layers, timesteps, or modules, making it difficult to compose interventions across the model.

In contrast, large language models benefit from a mature ecosystem of causal analysis tools operating on core architectural components such as residual streams, attention heads, and MLP blocks. Libraries such as TransformerLens \citep{transformerlens} and NNsight \citep{toolkit_nnsight} provide standardized abstractions for activation tracing and intervention, and methods including classifier-guided steering \citep{oozeer2025beyond}, representation fine-tuning \citep{wu2024reftrepresentationfinetuninglanguage}, and model-stitching analyses \citep{oozeeractivation, lee2025shared} are routinely applied across models and architectures.



Diffusion models therefore lack a model-agnostic abstraction that treats the diffusion stack as a structured target for causal analysis and intervention. To address this limitation, we introduce \textsc{DreamReader}, a framework unifying existing diffusion interpretability techniques within a single abstraction layer. \textsc{DreamReader} exposes the full T2I architecture, including text encoder, U-Net, denoising latents, and decoder, as configurable intervention sites. This abstraction allows us to derive three novel operators for diffusion models: LoReFT, classifier-guided gradient steering, and component-level cross-model mapping which were previously not applied to T2I models. Our contributions can be summarized as:
\begin{itemize}

\item \textbf{New methodological primitives for diffusion interpretability.}
We introduce three operators for T2I models:
(i) \emph{representation fine-tuning (LoReFT)} for subspace-constrained adaptation,
(ii) \emph{classifier-guided gradient steering}, which applies gradients of an MLP classifier trained on activations to induce targeted directional interventions, and
(iii) \emph{component-level cross-model mapping} via an affine transformation learned through ridge regression to align representations across heterogeneous diffusion modules. We conduct empirical case studies that show that these new methods yield promising and controllable interventions
in diffusion models.

\item \textbf{Empirical studies of stitching and steering.}
We demonstrate DreamReader with two case studies (Section~4). First, we transfer steering from a fine-tuned model to its base model by learning a lightweight MLP that maps activations between the two T2I models, enabling cross-model steering transfer with meaningful semantic edits. Second, we apply LoREFT steering on SDXL-Turbo U-Net block activations, showing effective control over generated images and providing evidence that LoREFT is a promising approach for image steering.

\item \textbf{A unified abstraction of the diffusion stack.}
DreamReader models text encoders, U-Nets, denoising latents, and decoders as structured intervention targets within a common framework (Table~\ref{tab:workflow-comparison}). This design supports coordinated interventions across timesteps and modules and eases comparison across architectures.

\item \textbf{Declarative and reproducible experimentation.}
DreamReader provides Hydra-style \citep{Yadan2019Hydra, hydra_rl} declarative experiment configurations specifying prompts, intervention sites, hyperparameters, and evaluation metrics (Figure~\ref{fig:t2i-interp-workflows}). The framework integrates reporting, artifact logging, and standardized metric computation for controlled sweeps and reproducible diffusion studies.

\end{itemize}

Together, these contributions establish a structured foundation for systematic causal analysis in text-to-image diffusion models.

\section{Toolkit Overview}

The toolkit is organized into four independent \emph{workflows} and a pluggable \emph{Reporting} module, as shown in Figure \ref{fig:t2i-interp-workflows}:
(1) \textbf{Localization}, (2) \textbf{Steering}, (3) \textbf{Model Stitching},
(4) \textbf{Sparse Autoencoders (SAEs)}, as well as a unified \textbf{Reporting} layer placed across the four workflows.
Each workflow exposes clear configs, inputs/outputs, standardized metrics, and cached artifacts for end-to-end reproducibility.

\paragraph{Execution model.}
Experiments are expressed through two abstract interfaces: \emph{Training}
(e.g., SAE training, linear probes, cross-model mappers) and \emph{Inference}
(e.g., attention-head interventions, SAE latent edits, model stitching).
Workflows are unified via Hydra~\cite{Yadan2019Hydra}, which provides
configuration-driven launches and native support for multi-run sweeps.
Each workflow is associated with a set of YAML configuration files that specify
(i) workflow- and function-specific inputs, (ii) dataset settings, and (iii)
reporting/logging details. Users typically edit a single configuration and
launch one or more jobs through the Hydra API; Hydra callbacks can be used to
aggregate results and produce comparative reports across runs. Jobs often
include hyperparameter searches, which can be invoked through Hydra sweeps; for
convenience, the same workflows can also be launched via provided bash scripts.


\paragraph{Reproducibility.}
Every run logs a fingerprint consisting of the model identifier,
dataset version, random seeds, train/eval splits, and
full intervention specifications (hook sites, features, scales, schedules).
Artifacts such as images and trained mappers may be
materialized to disk or exported to W\&B~\footnote{\url{https://wandb.ai/}}.

\paragraph{Reporting.}
The Reporting module consumes workflow artifacts and emits standardized
summaries (YAML) as well as optional W\&B or PDF reports. Reporting is decoupled, so new interfaces and UIs can be added or reused across projects without modifying core workflows.

We compare \textsc{DreamReader} against two of the most popular mechanistic interpretability libraries (TransformerLens and NNsight) in Table \ref{tab:toolkit-comparison}. 



\section{Workflows}


\label{sec:workflows}

\subsection{Localization}
Recent work on T2I models increasingly adapts causal tracing and ablation techniques from language models. One line of work focuses on \emph{token grounding}, identifying where individual prompt tokens are realized in the generated image \cite{tang2022daam}. A complementary line of work applies causal patching and ablations to U-Net representations (for example, modifying selected cross-attention heads or layers) to study component roles and assess whether specific objects, attributes, or styles can be selectively removed \citep{ghandeharioun2024patchscopes, basu2024mechanistic}. \textsc{DreamReader} supports these experiments by allowing users to specify prompts, target modules, and intervention policies within a unified interface, while providing built-in attention extraction, causal patching, and logging of counterfactual generations.

\begin{figure}[h]
  \centering
  \setlength{\tabcolsep}{3pt} 
  \renewcommand{\arraystretch}{0} 

  \begin{tabular}{@{}cc@{}}
    \begin{subfigure}[t]{0.245\textwidth}
      \centering
      \IfFileExists{images/baseline_man.png}{%
        \includegraphics[width=\linewidth]{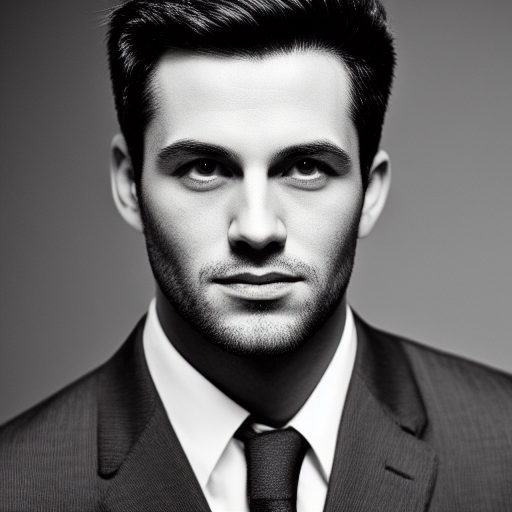}%
      }{%
        \fbox{\parbox[c][3.5cm][c]{0.9\linewidth}{\centering baseline\_man.png missing}}%
      }
      \caption{\textit{Baseline: photo of a man}}
      \label{fig:panel_tl}
    \end{subfigure} &
    \begin{subfigure}[t]{0.245\textwidth}
      \centering
      \IfFileExists{images/steered_black_man.png}{%
        \includegraphics[width=\linewidth]{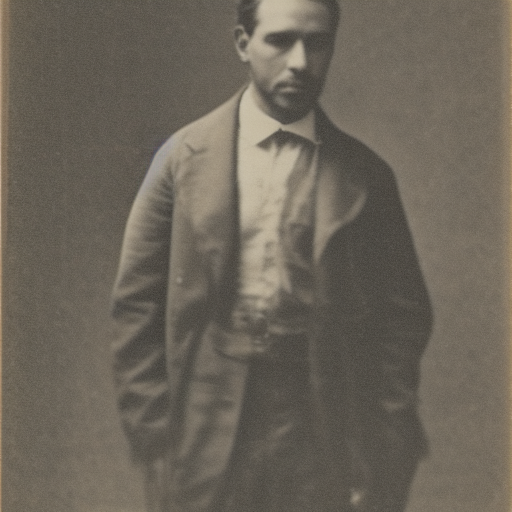}%
      }{%
        \fbox{\parbox[c][3.5cm][c]{0.9\linewidth}{\centering steered\_black\_man.png missing}}%
      }
      \caption{\textit{Steered: toward Black}}
      \label{fig:panel_tr}
    \end{subfigure} \\

    \begin{subfigure}[t]{0.245\textwidth}
      \centering
      \IfFileExists{images/baseline_child.png}{%
        \includegraphics[width=\linewidth]{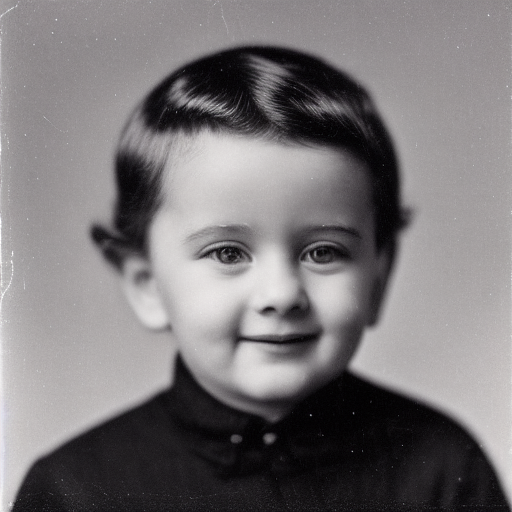}%
      }{%
        \fbox{\parbox[c][3.5cm][c]{0.9\linewidth}{\centering baseline\_child.png missing}}%
      }
      \caption{\textit{Baseline: photo of a child}}
      \label{fig:panel_bl}
    \end{subfigure} &
    \begin{subfigure}[t]{0.245\textwidth}
      \centering
      \IfFileExists{images/steered_sad_child.png}{%
        \includegraphics[width=\linewidth]{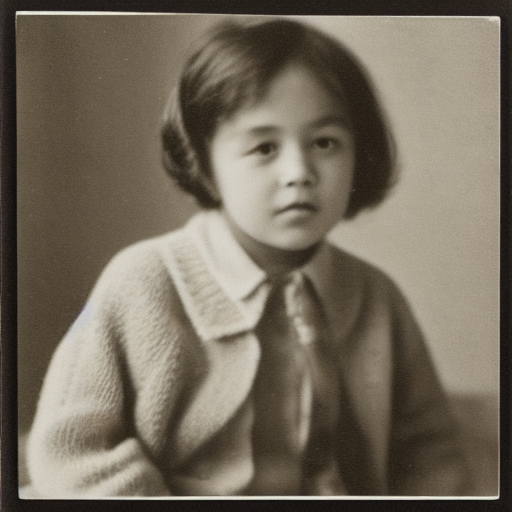}%
      }{%
        \fbox{\parbox[c][3.5cm][c]{0.9\linewidth}{\centering steered\_sad\_child.png missing}}%
      }
      \caption{\textit{Steered: toward sad}}
      \label{fig:panel_br}
    \end{subfigure}
  \end{tabular}

  \caption{\textbf{Steering.} We use CAA to extract a steering direction from a fine-tuned SD~1.5 model using a minimal contrastive setup with two prompts (e.g., \textit{``a photo of a Black man''} vs.\ \textit{``a photo of a man''}). We map this direction to the base model with our learned mapper and apply it during generation. The figure compares baseline samples (left) with the corresponding steered samples (right) for two example prompts.}
  \label{fig:stitching_results}
\end{figure}
\subsection{Steering (targeted interventions on T2I models)}
Beyond localization, a growing line of work studies \emph{steering} T2I models by editing internal activations along learned concept directions, for style adaptation \cite{gandikota2024concept}, concept erasure \cite{gaintseva2025casteer}, or attribute control \cite{li2024self}. CASteer \citep{gaintseva2025casteer} estimates concept vectors in cross-attention space from positive/negative prompt pairs and then adds or subtracts these directions during denoising to strengthen or suppress harmful attributes. Activation steering is also being explored for de-biasing T2I models \citep{kim2025rethinking,zhang2023iti,chinchure2023tibet}.

\textsc{DreamReader} provides a unified API for these techniques and extends them by introducing steering methods originally developed for language models, including K-Steer \cite{oozeer2025beyond}, and LoReFT \cite{wu2024reftrepresentationfinetuninglanguage}, into the diffusion setting. Given a set of modules (e.g., cross-attention or text encoder blocks), users can estimate steering directions and apply them at test time across a range of parameters to add, remove, or rescale concepts in diffusion trajectories. This enables systematic comparison of steering methods at the level of individual concepts.


\begin{figure}[t]
\centering
\begin{minipage}{0.98\linewidth}
\begin{lstlisting}
# (1) Initialize workflow object and mapper
stitcher = Stitcher()
input_dim  = 77 * 768
hidden_dim = 11 * 768
output_dim = 4 * 64 * 64
mapper = MLPMapper(input_dim=input_dim, output_dim=output_dim, hidden_dim=hidden_dim)

# (2) Define training specification (data, optimization, logging, precision)
spec = TrainingSpec(
    training_function=stitcher.train_mapper,
    kwargs={
        "train_loader"    : train_loader,
        "val_loader"      : val_loader,
        "mapper"          : mapper,
        "optimizers"      : [th.optim.Adam(mapper.parameters(), lr=1e-4)],
        "num_steps"       : 1_000,
        "loss_fn"         : th.nn.MSELoss(),
        "training_device" : "cuda:0",
        "autocast_dtype"  : th.bfloat16,
        "log_steps"       : 100,
    }
)

# (3) Run training and retrieve the trained mapper
output = Training(spec).run_trainer()
trained_mapper = output.preds  # trained mapper returned by the trainer
print("Training complete. Mapper:", trained_mapper)
\end{lstlisting}
\end{minipage}
\caption{\textbf{Stitching workflow (code snippet).} Example usage of the \textsc{DreamReader} \texttt{Stitcher} to train an \texttt{MLPMapper} that maps source-model activations to a target activation space. The snippet instantiates the workflow and mapper (1), specifies optimization, precision, and logging via \texttt{TrainingSpec} (2), and executes \texttt{run\_trainer()} to obtain the trained mapper from \texttt{output.preds} (3).}
\label{fig:stitching-workflow-snippet}
\end{figure}
\subsection{Stitching}

A third component of our toolkit is \emph{stitching}, which includes skipping modules in the model or learning mappings between internal representations of different models or between distinct modules of the same model (e.g., from a text encoder block into a U-Net block). This is inspired by representation-similarity and model-stitching studies \citep{Lenc2014UnderstandingIR, gandikota2025distilling} showing that when two networks learn compatible feature spaces, a simple affine or shallow MLP adapter can be trained to transfer activations from one into the other while preserving task performance. The workflow takes two modules (\texttt{module\_a} and \texttt{module\_b}) and their corresponding models as input, along with training parameters for the adapter, as illustrated in Figure~\ref{fig:stitching-workflow-snippet}.

\begin{table*}[t]
\centering
\small
\begin{tabularx}{\textwidth}{lXXX}
\toprule
\textbf{Capability} 
  & \textbf{\textsc{Dream Reader} (Ours)} 
  & \textbf{TransformerLens} 
  & \textbf{nnsight} \\
\midrule
Primary focus 
  & Diffusion / T2I 
  & Autoregressive LLMs 
  & LLMs (+ diffusion) \\

Abstraction level 
  & End-to-end workflows 
  & Mechanistic primitives 
  & Tracing/editing primitives \\

Tracing \& editing 
  & (steer, ablate, stitch) 
  & Patch-only 
  & Patch + causal \\

SAE support 
  & Built-in workflow 
  & External add-ons 
  & External add-ons \\

Experiment infrastructure 
  & YAML pipelines + standardized outputs 
  & Python scripts 
  & Python scripts \\

Visualization \& reporting 
  & Integrated tables 
  & External tooling 
  & External tooling \\
\bottomrule
\end{tabularx}

\caption{
Comparison of \textsc{Dream Reader} with existing interpretability libraries. 
“Tracing \& editing” encompasses activation extraction, intervention (e.g., patching, steering, ablations), and model stitching. 
“Experiment infrastructure” refers to configuration-driven experiment setup and standardized output formats. 
TransformerLens focuses on low-level mechanistic analysis for language models, while nnsight provides a general tracing and intervention graph framework. 
\textsc{Dream Reader} differs by providing diffusion-first, workflow-oriented pipelines with integrated experiment management and reporting.
}
\label{tab:toolkit-comparison}
\end{table*}

\subsection{Sparse Autoencoders (SAEs)}
\label{subsec:saes}

Sparse autoencoders (SAEs) decompose dense activations into a compact basis whose latents align with meaningful visual patterns, enabling targeted editing and concept discovery in T2I models \citep{surkov2024one}. The SAE workflow has two stages: (i) training, supporting variants such as Top-$k$, Gated, and Matryoshka~\citep{matryoshka,cunningham2023sparse,gated}, and (ii) intervention through ablation or scaling of learned latents.

\section{Case Studies}

\paragraph{Stitching.}

We perform model stitching in activation space by learning an MLP mapper between the UNet \texttt{mid\_block} activations of a LoRA-fine-tuned SD~1.5 model and the SD~1.5 base model, with all inputs (HuggingFace model identifiers, dataset and column, hooked \texttt{torch.nn} modules, denoising timestep(s) for activation collection, and mapper architecture/hyperparameters) specified in a Hydra config. The workflow supports two modes: in \texttt{MODE=train}, we collect paired activations from both models on the specified modules and train the user-chosen mapper to predict base-model activations from fine-tuned activations, saving the mapper checkpoint for reuse; in \texttt{MODE=steer}, we apply a contrastive steering direction defined by prompt triples of the form \texttt{prompt\textsubscript{+}$|$prompt\textsubscript{-}$|$prompt\textsubscript{eval}} and map this direction into the base model using the learned mapper. Steering is launched as a Hydra multirun (\texttt{-m}) over multiple prompt triples (e.g., \texttt{STEER\_PAIRS\_STR=``a photo of a Black man|a photo of a man|a photo of a man"}), and the outputs from all jobs are automatically aggregated into a single report (image grids and summary statistics), saved to disk and optionally uploaded to Weights \& Biases. The prompt-triple template is included only as a simple demonstration; in general, steering vectors can be obtained from datasets using CAA or other techniques, and our pipeline is agnostic to the choice of steering method. Figure~\ref{fig:stitching_results} shows example results from two such steering jobs.

\paragraph{Concept Addition}

We apply LoREFT~\cite{wu2024reftrepresentationfinetuninglanguage} as a lightweight mechanism for steering semantic concepts in T2I models. As a proof of concept, we target the attribute \emph{spectacles} in SDXL-Turbo. We construct a minimal paired prompt set (\(\sim\)1k examples) with and without spectacles, and train LoREFT to learn a low-rank subspace intervention on the UNet cross-attention blocks (down/mid/up) across all denoising steps (4). At inference time, we apply the learned intervention to a small set of held-out prompts and visualize the resulting generations in Figure~\ref{fig:loreft_result}. While LoREFT induces spectacles reliably for some prompts, other cases remain challenging, motivating a broader sweep over layers, step schedules, and training data in future work.

\section{Related Work in Interpretability Toolkits}
For vision and multimodal models, prior toolkits are relatively newer and less unified than those for language models. Prisma \cite{joseph2025prismaopensourcetoolkit} targets mechanistic interpretability of vision and video models, filling a role analogous to TransformerLens for vision. In the T2I setting, practitioners often rely on NNsight as a general intervention backend.
 Sparse autoencoders form a complementary pillar of this ecosystem. Libraries such as dictionary\_learning \cite{marks2024dictionarylearning}, Overcomplete \cite{fel2025overcomplete}, and Sparsify \cite{eleutherai_sparsify_2025} provide utilities for training $k$-sparse autoencoders on activations from language and vision models but an equivalent library for T2I models is lacking. SAE-Lens focuses on training and analysing SAEs specifically on LLM activations. 
Finally, CircuitsVis \cite{cooney2023circuitsvis} offers reusable visualization components that make circuit-level analyses easier to conduct. Diffusion models have no such toolkits, however refer to Section \ref{sec:summary_diffusion_interp} for an overview of interpretability techniques introduced in the field, and how \textsc{DreamReader} compares.

\section{Conclusion}

We presented \textsc{DreamReader}, a framework for systematic intervention in text-to-image diffusion models. By unifying existing techniques and introducing new operators such as LoReFT, classifier-guided gradient steering, and affine cross-model mapping, it expands the methodological toolkit available for studying and controlling diffusion systems. Our case studies on steering and model stitching illustrate how these tools enable controlled interventions across timesteps and attention components. A declarative YAML configuration layer, together with structured logs and artifacts, supports reproducible and extensible experimentation.
Overall, \textsc{DreamReader} provides a practical foundation for deeper investigation of representation structure in generative diffusion models.
\section{Limitations and Future Work}
In future iterations, we plan to extend \textsc{DreamReader} beyond current diffusion backbones to support Flux-style text-to-image models as well as vision–language models (VLMs), enabling unified workflows across architectures. We also plan to integrate \textsc{DreamReader} with Neuronpedia, enabling artifacts such as trained SAEs and steering vectors to be published in a unified format and easily shared with the wider community.

\section*{Ethics Statement}

Our work introduces a toolkit for text-to-image diffusion models with the goal of increasing transparency, auditability, and controllability of generative systems, especially in the context of bias, safety, and downstream societal impact. At the same time, such tooling is inherently dual-use: the same mechanisms that help localize and reduce biased behavior could, in principle, be repurposed to amplify harmful features, generate more targeted disallowed content, or construct stronger jailbreaks and adversarial interventions, and we explicitly document this dual-use nature in the toolkit README and documentation. Our analyses of demographic attributes (e.g., race, gender, age) in generated images are conducted with the explicit aim of measuring and reducing representational harms; we caution against using our toolkit to make normative claims about individuals or groups. Finally, interpretability results are approximate, localized “features’’ may not fully capture the causal structure of model behavior, and steering or ablating features can have unintended side effects or fail to generalize beyond specific benchmarks; we therefore discourage using the toolkit as the sole basis for high-stakes deployment decisions and instead envision it as a research and auditing tool that complements broader safety, governance, and human-in-the-loop evaluation practices.

\appendix
\section{Appendix}
\label{sec:appendix}

\begin{table*}[t]
\centering
\small
\renewcommand{\arraystretch}{1.02}
\setlength{\tabcolsep}{2.8pt}

\begin{tabular}{p{0.16\textwidth} *{9}{c}}
\toprule
\multirow{2}{*}{\textbf{System / Paper}} &
\multicolumn{6}{c}{\textbf{Interpretability Capabilities}} &
\multicolumn{3}{c}{\textbf{Visualization \& Workflow}} \\
\cmidrule(lr){2-7}\cmidrule(lr){8-10}
& \begin{tabular}{@{}c@{}}Head/Block\\Patching\end{tabular}
& \begin{tabular}{@{}c@{}}Concept\\Directions\end{tabular}
& \begin{tabular}{@{}c@{}}SAE\\Training\end{tabular}
& \begin{tabular}{@{}c@{}}SAE\\Intervene\end{tabular}
& \begin{tabular}{@{}c@{}}Model\\Stitching\end{tabular}
& \begin{tabular}{@{}c@{}}Trained\\Steering\\(LoREFT/K-Steer)\end{tabular} 
& \begin{tabular}{@{}c@{}}Attention\\Maps\end{tabular}
& \begin{tabular}{@{}c@{}}Intermed.~Viz\\($\hat{x}_0$/steps)\end{tabular}
& \begin{tabular}{@{}c@{}}Workflow\\\& Reports\end{tabular} \\
\midrule

\textit{Patch Explorer} \citep{grabe2025patch}
& \ck & \na & \na & \na & \na & \na & \parti & \parti & \na \\

\textit{One-Step is Enough (SAE)} \citep{surkov2024one}
& \na & \na & \ck & \ck & \na & \na & \na & \na & \parti \\

\textit{Cross/Self-Attn for Editing} \citep{liu2024towards}
& \na & \na & \na & \na & \na & \na & \ck & \na & \na \\

\textit{Distilling Diversity \& Control} \citep{gandikota2025distilling}
& \na & \na & \na & \na & \parti & \na & \parti & \ck & \parti \\

\textit{SliderSpace} \citep{gandikota2025sliderspace}
& \na & \ck & \na & \parti & \na & \na & \na & \ck & \parti \\

\textit{DAAM} \citep{tang2022daam}
& \na & \na & \na & \na & \na & \na & \ck & \na & \na \\

\textit{InterpretDiffusion} \citep{li2024self}
& \na & \ck & \na & \parti & \na & \na & \na & \na & \na \\

\textit{CASteer} \citep{gaintseva2025casteer}
& \na & \ck & \na & \ck & \na & \na & \na & \na & \parti \\

\midrule
\textbf{\textsc{DreamReader} (ours)}
& \ck & \ck & \ck & \ck & \ck & \ck & \ck & \ck & \ck \\

\bottomrule
\end{tabular}

\caption{T2I interpretability repositories summary. Left block: core interpretability capabilities; right block: visualization and workflow support. Symbols: \ck supported; \parti partial/limited; \na not reported/applicable.}
\label{tab:t2i_unified_comparison}
\end{table*}

\label{app:grouping}

\subsection{Summarizing Diffusion Interpretability Methods}
\label{sec:summary_diffusion_interp}
For input attribution and localization, cross-attention aggregation yields token-to-pixel maps that align with human judgments and expose co-hyponym or adjectival entanglement \citep{tang2022daam}, while more black-box variants localize token information via mutual information between token embeddings and pixels using only denoiser outputs \citep{kong2023interpretable}. For probing and internal-state visualization, logit-lens-style reconstructions over diffusion steps determine key component and directions in the activation space \citep{gandikota2025sliderspace, toker2024diffusion}. For interventions and component-level editing, head- or block-level patching tools isualize and manipulate a component’s contribution to the generated image \citep{grabe2025patch}, while complementary component-attribution methods uncover editable positive and negative contributions of specific units or channels \citep{nguyen2025unveiling}. For sparse feature discovery, sparse autoencoders (SAEs) trained on transformer blocks enable semantically targeted edits in image generation \citep{surkov2024one}. Finally, work on the “universality” of embeddings and cross-model activation mapping in language models \citep{platonic,oozeeractivation,bello2025linear} suggests promising directions in T2I interpretability tools, such as cross-model comparison and feature transfer. We recap and situate our work in Table \ref{tab:t2i_unified_comparison}.

\begin{table*}[h]
\centering
\small
\begin{tabularx}{\textwidth}{lX}
\toprule
\textbf{Term} & \textbf{Definition} \\
\midrule

\textbf{Scales} & Intervention strength parameters (e.g., how strongly activations are modified). \\

\textbf{Schedules} & Temporal specifications describing when an intervention is applied (e.g., which denoising steps/timesteps). \\

\textbf{Steering vector} & A learned direction in activation space that, when added to activations, steers model behavior. \\

\textbf{Mapper} & A neural network (affine/linear or MLP) transforming activations, used in steering and stitching workflows. \\

\textbf{Feature ablation} & Removing or zeroing specific features to test their causal role. \\

\textbf{CLIP alignment} & Semantic similarity between text prompt and generated image using CLIP embeddings. \\

\textbf{FID/KID} & Image distribution metrics: Fréchet Inception Distance and Kernel Inception Distance. \\

\textbf{LPIPS} & Learned Perceptual Image Patch Similarity; measures perceptual closeness. \\

\textbf{SSIM} & Structural Similarity Index; measures structural similarity across images. \\

\textbf{Artifacts} & Materialized outputs saved to disk (images, model checkpoints, trained modules). \\

\textbf{Reports} & Standardized summaries, plots, tables, and diagnostics derived from artifacts. \\

\bottomrule
\end{tabularx}
\caption{Glossary of key terms used across DreamReader workflows.}
\label{tab:workflow-glossary}
\end{table*}

\end{document}